\documentclass[conference]{IEEEtran}
\IEEEoverridecommandlockouts
\usepackage{cite}
\usepackage{mathtools}
\usepackage{amsmath,amssymb,amsfonts}
\usepackage{algorithmic}
\usepackage{hyperref}
\usepackage{graphicx}
\usepackage{textcomp}
\usepackage{xcolor}
\usepackage{fancyhdr}
\usepackage{float}
\def\BibTeX{{\rm B\kern-.05em{\sc i\kern-.025em b}\kern-.08em
    T\kern-.1667em\lower.7ex\hbox{E}\kern-.125emX}}

\begin{document}

\title{
Leveraging Generative AI Models for Synthetic Data Generation in Healthcare: Balancing Research and Privacy}

\author{
\IEEEauthorblockN{Aryan Jadon}
\IEEEauthorblockA{
\textit{
San Jose State University}\\
CA, USA \\
aryan.jadon@sjsu.edu
}

\and

\IEEEauthorblockN{Shashank Kumar}
\IEEEauthorblockA{\textit{University of Florida} \\
FL, USA \\
sh.kumar@ufl.edu}
}

\maketitle
\begin{abstract}
The widespread adoption of electronic health records and digital healthcare data has created a demand for data-driven insights to enhance patient outcomes, diagnostics, and treatments. However, using real patient data presents privacy and regulatory challenges, including compliance with HIPAA \cite{annas2003hipaa} and GDPR \cite{beaulieu2020privacy}. Synthetic data generation, using generative AI models like GANs \cite{goodfellow2014generative} and VAEs \cite{kingma2013auto}, offers a promising solution to balance valuable data access and patient privacy protection. In this paper, we examine generative AI models for creating realistic, anonymized patient data for research and training \cite{choi2017generating}, explore synthetic data applications in healthcare, and discuss its benefits, challenges, and future research directions. Synthetic data has the potential to revolutionize healthcare by providing anonymized patient data while preserving privacy and enabling versatile applications.

\end{abstract}

\begin{IEEEkeywords}
Generative AI Models, Synthetic Data Generation, Healthcare Research, Patient Privacy, Data Augmentation, Federated Learning, Differential Privacy, Data Anonymization, Transfer Learning, Health Informatics.
\end{IEEEkeywords}

\section{Introduction}

The rapid digitalization of healthcare data and the increasing adoption of electronic health records (EHRs) have opened new opportunities for leveraging data-driven insights to improve patient care, diagnostics, and treatment \cite{sevakula2020state}. However, the use of real patient data often raises concerns regarding privacy and compliance with data protection regulations such as the Health Insurance Portability and Accountability Act (HIPAA)\cite{annas2003hipaa} and the General Data Protection Regulation (GDPR) \cite{beaulieu2020privacy}. These concerns create barriers for researchers and healthcare professionals who require access to large amounts of data to develop and validate advanced algorithms and AI models\cite{jadon2023overview}.

Synthetic data generation, powered by generative AI models, offers a promising solution to balance the need for data-driven insights with patient privacy. Generative AI models, such as Generative Adversarial Networks (GANs) \cite{goodfellow2014generative} and Variational Autoencoders (VAEs) \cite{kingma2013auto}, learn the underlying structure and distribution of real-world data to generate new, synthetic instances with similar characteristics. By creating realistic, anonymized patient data, these models ensure that sensitive patient information is protected while providing researchers with valuable data for analysis and training purposes \cite{choi2017generating}.

\begin{figure}[htbp]
\centerline{\includegraphics[width=8cm]{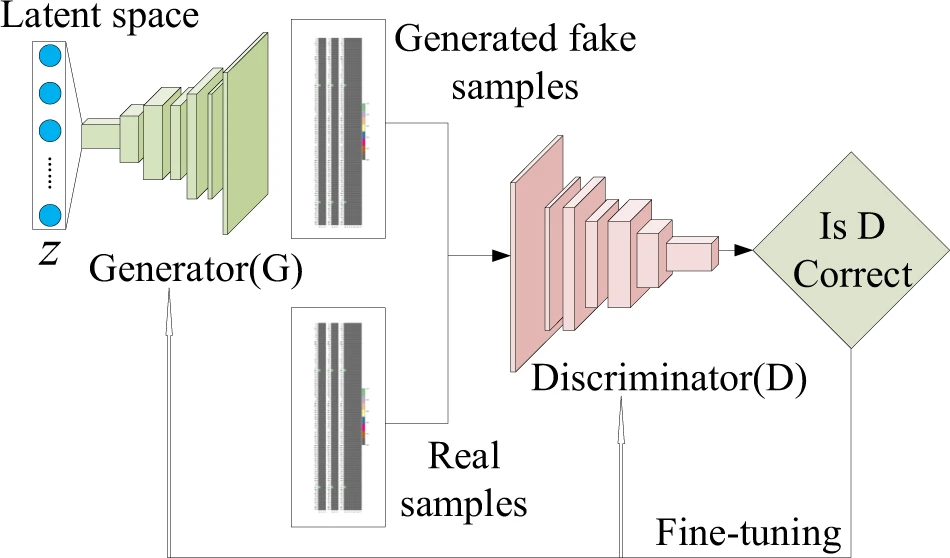}}
\caption{Generative adversarial networks (GAN) based efficient sampling \cite{dan2020generative}}
\label{fig}
\end{figure}

In this paper, we explore the role of generative AI models in generating synthetic patient data and discuss their potential applications, benefits, and challenges in the healthcare domain. We also present future research directions and the potential impact of synthetic data on healthcare research and practice, including its integration with other privacy-preserving techniques such as differential privacy \cite{abadi2016deep}, federated learning \cite{rieke2020future} and data anonymization\cite{mohammed2010centralized}. This comprehensive examination of synthetic data generation using generative AI models aims to provide a foundation for understanding the value of this approach in healthcare and its potential to revolutionize research, diagnostics, and treatment while safeguarding patient privacy. 
Our Github Repo can be found at \href{https://github.com/aryan-jadon/Synthetic-Data-Medical-Generative-AI}{https://github.com/aryan-jadon/Synthetic-Data-Medical-Generative-AI}.

\section{Generative AI Models for Synthetic Data Generation}
Generative AI models, which include Generative Adversarial Networks (GANs) \cite{goodfellow2014generative} and Variational Autoencoders (VAEs) \cite{kingma2013auto}, are designed to learn the underlying structure and distribution of real-world data and generate new, synthetic instances with similar characteristics. These models have shown remarkable success in generating high-quality synthetic data across various domains, including healthcare.

\begin{enumerate}
\item \textbf{Generative Adversarial Networks (GANs):} GANs consist of two neural networks, a generator, and a discriminator, that compete with each other in a minimax game. The generator creates synthetic data samples, while the discriminator distinguishes between real and generated samples. The generator improves its output by attempting to deceive the discriminator, resulting in increasingly realistic synthetic data.

\begin{figure}[htbp]
\centerline{\includegraphics[width=8cm]{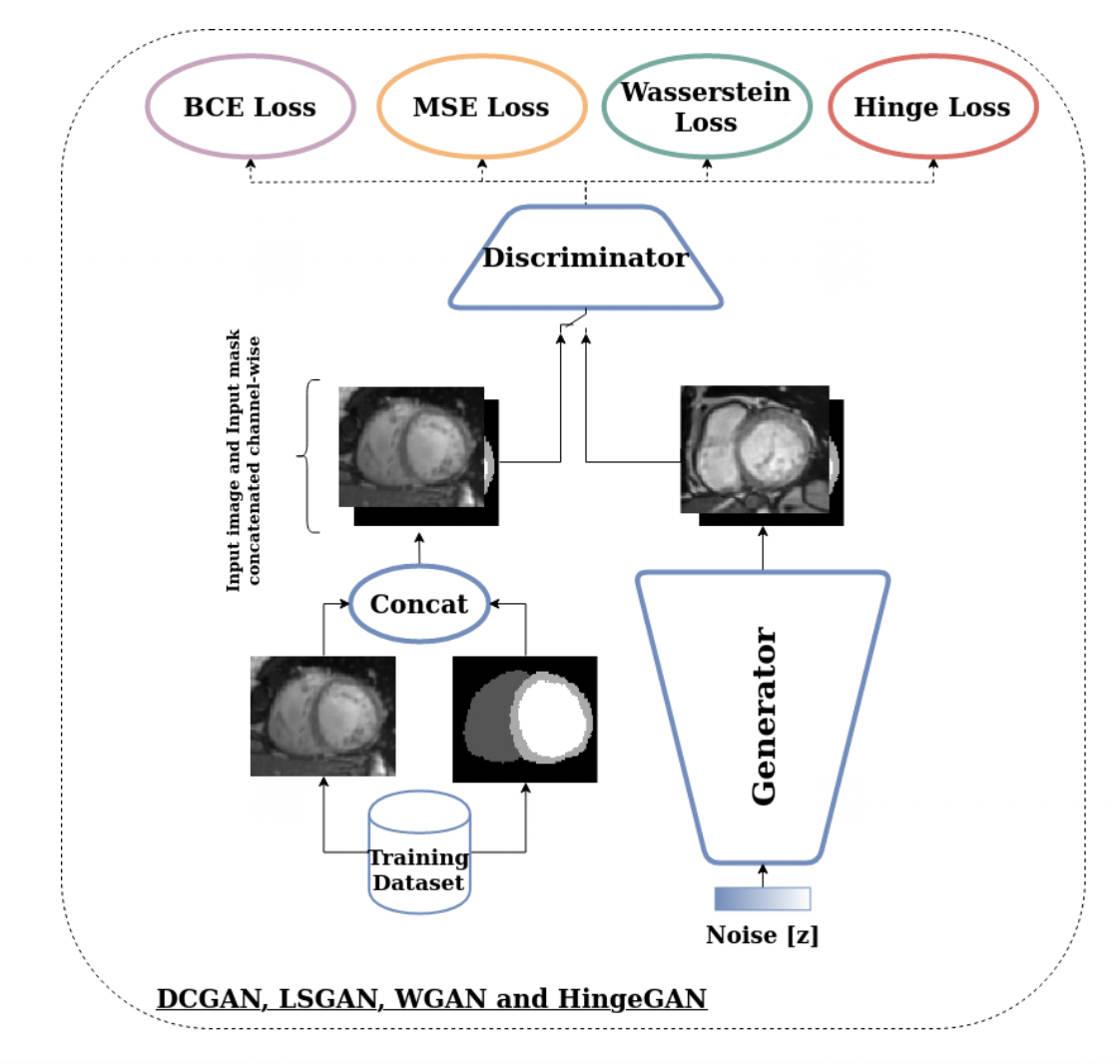}}
\caption{Representative Architecture of wide variety of GAN's \cite{sk2021gans}}
\label{fig}
\end{figure}

\item \textbf{Variational Autoencoders (VAEs):} VAEs are a class of generative models that combine autoencoders with variational inference. They learn a probabilistic mapping between data and latent space, enabling the generation of new data samples by sampling from the latent space. VAEs have been used to generate realistic synthetic data while maintaining a balance between data fidelity and diversity.

\begin{figure}[htbp]
\centerline{\includegraphics[width=8cm]{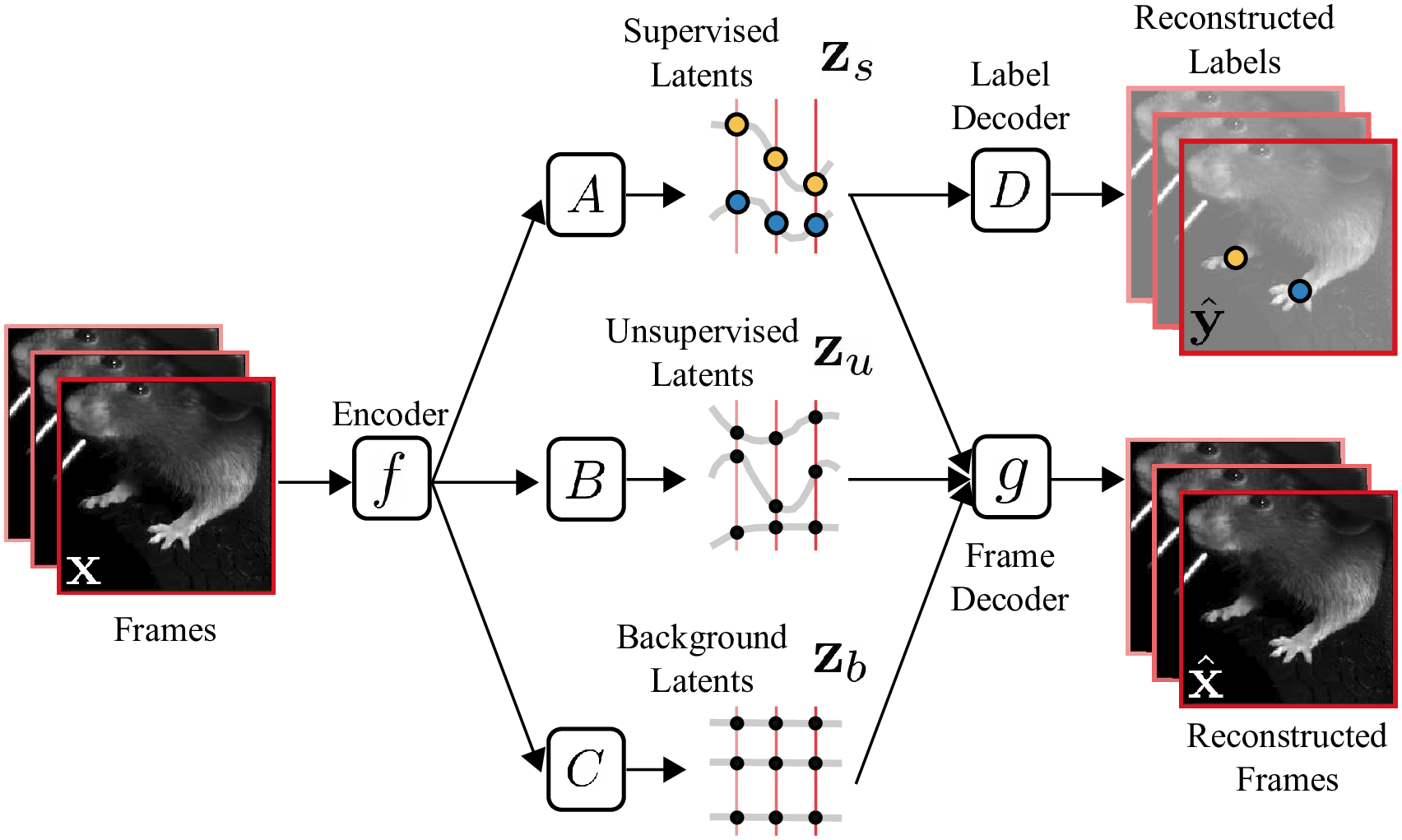}}
\caption{Application of Variational Autoencoders in Medical Experiments \cite{whiteway2021partitioning}}
\label{fig}
\end{figure}

\end{enumerate}

\subsection{Synthetic Data Generation Process}

The process of generating synthetic data using generative AI models involves three main steps:

\begin{enumerate}
\item \textbf{Training generative models on real-world data:} The model is trained using a dataset of real patient data, which allows it to learn the underlying structure, relationships, and distributions present in the data.
\item \textbf{Generating new, synthetic instances with similar characteristics:} Once trained, the generative model can create synthetic data samples that closely resemble the real data while preserving the underlying relationships and patterns. This process ensures that the generated data is both realistic and anonymized.

\item \textbf{Evaluating the quality and utility of synthetic data:} The generated synthetic data should be evaluated based on its resemblance to real data, its ability to maintain the underlying relationships and patterns, and its utility for the intended application, such as research or AI model training.
\end{enumerate}

By leveraging generative AI models for synthetic data generation, healthcare researchers and professionals can access realistic, anonymized patient data while addressing concerns related to patient privacy and regulatory compliance.




\section{Applications of Synthetic Data in Healthcare}


Synthetic data generated using generative AI models have a wide range of applications in healthcare \cite{dahmen2019synsys}, enabling researchers and professionals to access realistic, anonymized patient data while maintaining privacy and compliance with data protection regulations. Some of the key applications are:

\subsection{AI Model Training and Validation}

Access to large, diverse datasets is crucial for training and validating AI models in healthcare. Synthetic data provides an alternative to real patient data, enabling researchers to develop and evaluate algorithms without the risk of exposing sensitive information.

\begin{itemize}
\item \textbf{Data augmentation:} Synthetic data can be used to augment existing real-world datasets, particularly when the available data is scarce or imbalanced. This augmentation can improve the performance and generalizability of AI models across different patient populations and clinical scenarios \cite{frid2018gan}.
\item \textbf{Model validation:} By generating synthetic data that closely resembles real-world data, researchers can validate the performance of AI models and ensure that they are robust and reliable in real-world settings\cite{chen2019validity}.
\end{itemize}

\subsection{Simulations for Medical Training and Decision Support}

Synthetic data can be used to create realistic simulations for medical training and decision support systems, allowing healthcare professionals to practice and improve their skills without the risk of harming real patients.

\begin{itemize}
\item \textbf{Medical training:} Synthetic data can be used to develop virtual patient cases and scenarios for training medical students, residents, and other healthcare professionals. These simulations provide a safe and controlled environment for learning and practicing clinical skills, diagnosis, and treatment planning \cite{mendelevitch2021fidelity}.
\item \textbf{Decision support systems:} Synthetic data can be integrated into clinical decision support systems to provide real-time guidance and recommendations based on the analysis of anonymized patient data. This approach can help healthcare professionals make more informed and personalized treatment decisions, ultimately leading to better patient outcomes. \cite{cresswell2012computerised}
\end{itemize}

\subsection{Healthcare Research}

Synthetic data allows researchers to conduct large-scale studies and analyses without the need for accessing real patient data, reducing the risk of privacy breaches and ensuring compliance with data protection regulations.

\begin{itemize}
\item \textbf{Epidemiological studies:} Researchers can use synthetic data to study the distribution, determinants, and outcomes of health-related conditions and diseases. This approach can provide valuable insights into the risk factors and preventive measures for various health issues, ultimately informing public health policies and interventions \cite{bhagyashree2018diagnosis}.

\item \textbf{Clinical trials:} Synthetic data can be employed to simulate patient populations, treatment groups, and outcomes in clinical trials. This approach can help researchers optimize trial designs, estimate the potential efficacy and safety of interventions, and identify potential biases and confounders \cite{weissler2021role}.
\end{itemize}

By leveraging synthetic data in these various applications, healthcare researchers and professionals can gain valuable insights, develop advanced AI models, and improve clinical practice while safeguarding patient privacy and maintaining compliance with data protection regulations.

\section{ Benefits and Challenges of Synthetic Data in Healthcare}
The use of synthetic data in healthcare offers several benefits but also presents challenges that must be considered and addressed to ensure its effective and responsible application.

\subsection{Benefits}
\begin{enumerate}
\item \textbf{Privacy preservation:} Synthetic data helps protect patient privacy by generating anonymized data instances that closely resemble real-world data without exposing sensitive information. This approach ensures compliance with data protection regulations such as HIPAA\cite{cohen2018hipaa} and GDPR\cite{larrucea2020towards}.

\item \textbf{Cost and time savings:} Accessing and sharing real-world patient data often requires significant resources and time due to the need for data anonymization, consent management, and compliance with legal and ethical requirements. Synthetic data can alleviate these burdens, enabling researchers and professionals to access and share data more efficiently.

\item \textbf{Data utility and quality:} Synthetic data can maintain the underlying relationships and patterns present in real-world data, ensuring its utility and quality for research, AI model training, and other applications. In addition, synthetic data can be used for data augmentation to enhance the performance and generalizability of AI models.
\end{enumerate}

\subsection{Challenges}

\begin{enumerate}
\item \textbf{Maintaining data fidelity and diversity:} Generating synthetic data that accurately represents real-world data while preserving diversity is a complex task. Overfitting or generating unrealistic data may limit the utility of synthetic data for research and AI model training \cite{naeem2020reliable}.

\item \textbf{Potential biases:} Synthetic data generated from real-world data may inadvertently reproduce or amplify existing biases, which could negatively impact the fairness and effectiveness of AI models and research outcomes. It is crucial to identify and mitigate potential biases in both real and synthetic data to ensure equitable and accurate results \cite{gianfrancesco2018potential}.

\item \textbf{Model complexity and computational resources:} Generative AI models, such as GANs and VAEs, can be computationally expensive and complex, requiring substantial resources for training and optimization. Researchers and professionals must carefully consider the trade-offs between model complexity, data quality, and computational resources.

\end{enumerate}

By addressing these challenges and harnessing the benefits of synthetic data, healthcare researchers and professionals can effectively use generative AI models to access realistic, anonymized patient data, facilitating advances in research, diagnostics, and treatment while maintaining patient privacy and compliance with data protection regulations.

\section{Future Research Directions and Impact}

There are several promising future research directions and potential impacts of synthetic data generation in healthcare:

\begin{enumerate}
\item \textbf{Advances in generative AI models:} As generative AI models continue to evolve, improvements in the quality, diversity, and fidelity of synthetic data can be expected. This will enable more accurate, versatile applications in healthcare research, AI model training, and clinical practice.

\item \textbf{Integration with privacy-preserving techniques:} Combining synthetic data generation with other privacy-preserving techniques, such as differential privacy and federated learning, can further enhance the privacy and utility of data for healthcare applications while minimizing the risk of privacy breaches.

\item \textbf{Expanding applications in healthcare:} Synthetic data generation can be applied to a wide range of healthcare domains, including personalized medicine, telemedicine, and public health surveillance. By providing access to realistic, anonymized patient data, synthetic data can help accelerate research and improve patient outcomes in these areas.
\end{enumerate}

The successful development and application of synthetic data generation in healthcare have the potential to revolutionize the way researchers and professionals access and use patient data, ultimately leading to significant advancements in diagnostics, treatment, and overall patient care, while safeguarding privacy and ensuring compliance with data protection regulations.

\section{Conclusion}
In this paper, we explored the role of generative AI models, such as GANs and VAEs, in generating realistic, anonymized synthetic patient data for research and training purposes in healthcare. By addressing the challenges associated with privacy and regulatory compliance, synthetic data can facilitate advancements in AI model development, medical training, healthcare research, and decision support systems. As generative AI models continue to evolve, future research directions include improving the quality and diversity of synthetic data, integrating privacy-preserving techniques, and expanding applications across various healthcare domains.

The potential impact of synthetic data generation in healthcare is immense, with the capability to revolutionize research, diagnostics, and treatment while maintaining patient privacy and compliance with data protection regulations. The successful application of synthetic data can ultimately lead to improved patient outcomes, more efficient healthcare systems, and a better understanding of the complex factors that influence human health.

\bibliographystyle{IEEEtran}
\nocite{*}
\bibliography{references}

\begin{thebibliography}{10}
\providecommand{\url}[1]{#1}
\csname url@samestyle\endcsname
\providecommand{\newblock}{\relax}
\providecommand{\bibinfo}[2]{#2}
\providecommand{\BIBentrySTDinterwordspacing}{\spaceskip=0pt\relax}
\providecommand{\BIBentryALTinterwordstretchfactor}{4}
\providecommand{\BIBentryALTinterwordspacing}{\spaceskip=\fontdimen2\font plus
\BIBentryALTinterwordstretchfactor\fontdimen3\font minus
  \fontdimen4\font\relax}
\providecommand{\BIBforeignlanguage}[2]{{%
\expandafter\ifx\csname l@#1\endcsname\relax
\typeout{** WARNING: IEEEtran.bst: No hyphenation pattern has been}%
\typeout{** loaded for the language `#1'. Using the pattern for}%
\typeout{** the default language instead.}%
\else
\language=\csname l@#1\endcsname
\fi
#2}}
\providecommand{\BIBdecl}{\relax}
\BIBdecl

\bibitem{annas2003hipaa}
G.~J. Annas, ``Hipaa regulations: a new era of medical-record privacy?''
  \emph{New England Journal of Medicine}, vol. 348, p. 1486, 2003.

\bibitem{beaulieu2020privacy}
B.~K. Beaulieu-Jones, W.~Yuan, S.~G. Finlayson, Z.~Wu, P.~Avillach, and I.~S.
  Kohane, ``Privacy-preserving generative deep neural networks support clinical
  data sharing,'' \emph{Circulation: Cardiovascular Quality and Outcomes},
  vol.~13, no.~7, p. e006246, 2020.

\bibitem{goodfellow2014generative}
I.~Goodfellow, J.~Pouget-Abadie, M.~Mirza, B.~Xu, D.~Warde-Farley, S.~Ozair,
  A.~Courville, and Y.~Bengio, ``Generative adversarial networks,'' \emph{arXiv
  preprint arXiv:1406.2661}, 2014.

\bibitem{kingma2013auto}
D.~P. Kingma and M.~Welling, ``Auto-encoding variational bayes,'' \emph{arXiv
  preprint arXiv:1312.6114}, 2013.

\bibitem{choi2017generating}
E.~Choi, S.~Biswal, B.~Malin, J.~Duke, W.~F. Stewart, and J.~Sun, ``Generating
  multi-label discrete patient records using generative adversarial networks,''
  \emph{arXiv preprint arXiv:1703.06490}, 2017.

\bibitem{sevakula2020state}
R.~K. Sevakula, W.-T.~M. Au-Yeung, J.~P. Singh, E.~K. Heist, E.~M. Isselbacher,
  and A.~A. Armoundas, ``State-of-the-art machine learning techniques aiming to
  improve patient outcomes pertaining to the cardiovascular system,''
  \emph{Journal of the American Heart Association}, vol.~9, no.~4, p. e013924,
  2020.

\bibitem{jadon2023overview}
S.~Jadon and A.~Jadon, ``An overview of deep learning architectures in few-shot
  learning domain,'' 2023.

\bibitem{dan2020generative}
Y.~Dan, Y.~Zhao, X.~Li, S.~Li, M.~Hu, and J.~Hu, ``Generative adversarial
  networks (gan) based efficient sampling of chemical composition space for
  inverse design of inorganic materials,'' \emph{npj Computational Materials},
  vol.~6, no.~1, p.~84, 2020.

\bibitem{abadi2016deep}
M.~Abadi, A.~Chu, I.~Goodfellow, H.~B. McMahan, I.~Mironov, K.~Talwar, and
  L.~Zhang, ``Deep learning with differential privacy,'' in \emph{Proceedings
  of the 2016 ACM SIGSAC conference on computer and communications security},
  2016, pp. 308--318.

\bibitem{rieke2020future}
N.~Rieke, J.~Hancox, W.~Li, F.~Milletari, H.~R. Roth, S.~Albarqouni, S.~Bakas,
  M.~N. Galtier, B.~A. Landman, K.~Maier-Hein \emph{et~al.}, ``The future of
  digital health with federated learning,'' \emph{NPJ digital medicine},
  vol.~3, no.~1, p. 119, 2020.

\bibitem{mohammed2010centralized}
N.~Mohammed, B.~C. Fung, P.~C. Hung, and C.-K. Lee, ``Centralized and
  distributed anonymization for high-dimensional healthcare data,'' \emph{ACM
  Transactions on Knowledge Discovery from Data (TKDD)}, vol.~4, no.~4, pp.
  1--33, 2010.

\bibitem{sk2021gans}
Y.~Skandarani, P.-M. Jodoin, and A.~Lalande, ``Gans for medical image
  synthesis: An empirical study,'' 2021.

\bibitem{whiteway2021partitioning}
M.~R. Whiteway, D.~Biderman, Y.~Friedman, M.~Dipoppa, E.~K. Buchanan, A.~Wu,
  J.~Zhou, N.~Bonacchi, N.~J. Miska, J.-P. Noel \emph{et~al.}, ``Partitioning
  variability in animal behavioral videos using semi-supervised variational
  autoencoders,'' \emph{PLoS computational biology}, vol.~17, no.~9, p.
  e1009439, 2021.

\bibitem{dahmen2019synsys}
J.~Dahmen and D.~Cook, ``Synsys: A synthetic data generation system for
  healthcare applications,'' \emph{Sensors}, vol.~19, no.~5, p. 1181, 2019.

\bibitem{frid2018gan}
M.~Frid-Adar, I.~Diamant, E.~Klang, M.~Amitai, J.~Goldberger, and H.~Greenspan,
  ``Gan-based synthetic medical image augmentation for increased cnn
  performance in liver lesion classification,'' \emph{Neurocomputing}, vol.
  321, pp. 321--331, 2018.

\bibitem{chen2019validity}
J.~Chen, D.~Chun, M.~Patel, E.~Chiang, and J.~James, ``The validity of
  synthetic clinical data: a validation study of a leading synthetic data
  generator (synthea) using clinical quality measures,'' \emph{BMC medical
  informatics and decision making}, vol.~19, no.~1, pp. 1--9, 2019.

\bibitem{mendelevitch2021fidelity}
O.~Mendelevitch and M.~D. Lesh, ``Fidelity and privacy of synthetic medical
  data,'' \emph{arXiv preprint arXiv:2101.08658}, 2021.

\bibitem{cresswell2012computerised}
K.~Cresswell, A.~Majeed, D.~W. Bates, and A.~Sheikh, ``Computerised decision
  support systems for healthcare professionals: an interpretative review.''
  \emph{Informatics in Primary Care}, vol.~20, no.~2, 2012.

\bibitem{bhagyashree2018diagnosis}
S.~I.~R. Bhagyashree, K.~Nagaraj, M.~Prince, C.~H. Fall, and M.~Krishna,
  ``Diagnosis of dementia by machine learning methods in epidemiological
  studies: a pilot exploratory study from south india,'' \emph{Social
  psychiatry and psychiatric epidemiology}, vol.~53, pp. 77--86, 2018.

\bibitem{weissler2021role}
E.~H. Weissler, T.~Naumann, T.~Andersson, R.~Ranganath, O.~Elemento, Y.~Luo,
  D.~F. Freitag, J.~Benoit, M.~C. Hughes, F.~Khan \emph{et~al.}, ``The role of
  machine learning in clinical research: transforming the future of evidence
  generation,'' \emph{Trials}, vol.~22, no.~1, pp. 1--15, 2021.

\bibitem{cohen2018hipaa}
I.~G. Cohen and M.~M. Mello, ``Hipaa and protecting health information in the
  21st century,'' \emph{Jama}, vol. 320, no.~3, pp. 231--232, 2018.

\bibitem{larrucea2020towards}
X.~Larrucea, M.~Moffie, S.~Asaf, and I.~Santamaria, ``Towards a gdpr compliant
  way to secure european cross border healthcare industry 4.0,'' \emph{Computer
  Standards \& Interfaces}, vol.~69, p. 103408, 2020.

\bibitem{naeem2020reliable}
M.~F. Naeem, S.~J. Oh, Y.~Uh, Y.~Choi, and J.~Yoo, ``Reliable fidelity and
  diversity metrics for generative models,'' in \emph{International Conference
  on Machine Learning}.\hskip 1em plus 0.5em minus 0.4em\relax PMLR, 2020, pp.
  7176--7185.

\bibitem{gianfrancesco2018potential}
M.~A. Gianfrancesco, S.~Tamang, J.~Yazdany, and G.~Schmajuk, ``Potential biases
  in machine learning algorithms using electronic health record data,''
  \emph{JAMA internal medicine}, vol. 178, no.~11, pp. 1544--1547, 2018.

\bibitem{voigt2017eu}
P.~Voigt and A.~Von~dem Bussche, ``The eu general data protection regulation
  (gdpr),'' \emph{A Practical Guide, 1st Ed., Cham: Springer International
  Publishing}, vol.~10, no. 3152676, pp. 10--5555, 2017.

\bibitem{rankin2020reliability}
D.~Rankin, M.~Black, R.~Bond, J.~Wallace, M.~Mulvenna, G.~Epelde \emph{et~al.},
  ``Reliability of supervised machine learning using synthetic data in health
  care: Model to preserve privacy for data sharing,'' \emph{JMIR medical
  informatics}, vol.~8, no.~7, p. e18910, 2020.

\end{thebibliography}
\end{document}